\documentclass[conference]{IEEEtran}
\usepackage{cite}
\usepackage{amsmath,amssymb,amsfonts}
\usepackage{algorithmic}
\usepackage{graphicx}
\usepackage{textcomp}
\usepackage{xcolor}
\def\BibTeX{{\rm B\kern-.05em{\sc i\kern-.025em b}\kern-.08em
    T\kern-.1667em\lower.7ex\hbox{E}\kern-.125emX}}
    
\begin{document}

\title{On the Use of Interpretable Machine Learning for the Management of Data Quality}


%
%
%

\author{\IEEEauthorblockN{Anna Karanika}
\IEEEauthorblockA{\textit{Dept. of Informatics and Telecommunications} \\
\textit{University of Thessaly}\\
Papasiopoulou 2-4, Lamia 35100 Greece  \\
ankaranika@uth.gr}
\and
\IEEEauthorblockN{Panagiotis Oikonomou}
\IEEEauthorblockA{\textit{Dept. of Informatics and Telecommunications} \\
\textit{University of Thessaly}\\
Papasiopoulou 2-4, Lamia 35100 Greece  \\
paikonom@uth.gr}
\and
\IEEEauthorblockN{Kostas Kolomvatsos}
\IEEEauthorblockA{\textit{Dept. of Informatics and Telecommunications} \\
\textit{University of Thessaly}\\
Papasiopoulou 2-4, Lamia 35100 Greece \\
kostasks@uth.gr}
\and
\IEEEauthorblockN{Christos Anagnostopoulos}
\IEEEauthorblockA{\textit{School of Computing Science} \\
\textit{University of Glasgow}\\
Lilybank Gardens 17, G12 8RZ Glasgow UK \\
christos.anagnostopoulos@glasgow.ac.uk}
}

\maketitle

\begin{abstract}
Data quality is a significant issue 
for any application that requests for analytics to support decision making.
It becomes very important when we focus on Internet of Things (IoT) 
where numerous devices can interact to exchange and process data.
IoT devices are connected to Edge Computing (EC) nodes to report the collected data, thus, we have to secure data quality not only at the IoT but also at the edge of the network.
In this paper, we focus on the specific problem and 
propose the use of interpretable machine learning to 
deliver the features that are important to be based for any 
data processing activity. 
Our aim is to secure data quality, at least, 
for those features that are detected as significant in the collected datasets.
We have to notice that the selected features depict the highest correlation with 
the remaining in every dataset, thus, they can be adopted for dimensionality reduction.
We focus on multiple methodologies for having interpretability in our learning models and adopt an ensemble scheme for the final decision.
Our scheme is capable of timely retrieving the final result and efficiently select the appropriate features.
We evaluate our model through extensive simulations and present numerical results.
Our aim is to reveal its performance under various experimental scenarios that we create varying a set of parameters adopted in our mechanism.  
\end{abstract}



\IEEEpeerreviewmaketitle

\section{Introduction}
Nowadays we are witnessing 
the advent of Internet of Things (IoT) where
numerous devices can interact with their environment and perform simple processing 
activities.
Multiple services and applications 
are executed over the humongous volumes of data 
collected by the IoT devices.
These data are transferred to the Cloud infrastructure to 
be the subject of further processing.
Due to network bandwidth, latency and
data privacy concerns, the research community has focused 
on the processing performed at the edge of the network.
Edge Computing (EC) involves heterogeneous nodes close to IoT
devices and end users capable of performing various 
activities and delivering analytics over the collected data.
EC nodes act as mediators between the IoT infrastructure and Cloud.
They can be sensors, home gateways, micro servers,
and small cells while being equipped with storage and computation
capabilities. 

Every EC node is `connected' to a number of IoT devices and 
become the host of the collected data.
We focus on a multivariate data scenario where multiple variables/dimensions/features
consist of vectors reported by IoT devices.
Locally, at EC nodes, an ecosystem of distributed datasets 
is formulated depicting the geo-located 
aspect of the problem. 
Data, before being the subject of processing, should be validated concerning their quality
to support efficient analytics.
A metric, among others, that secures data quality 
is accuracy \cite{loshin}.
Accuracy refers to the closeness of estimates 
to the (unknown) exact or true values \cite{refec}.
In other words, accuracy depicts the error between the 
observation/estimation and the real data.
We consider that maintaining accuracy in a dataset 
will lead to `solid' data repositories,
i.e., datasets exhibiting a limited error/deviation (around the mean).
Actually, `solid' datasets 
is the target of data separation algorithms 
proposed in the relevant literature;
these algorithms aim to deliver  
small non-overlapping datasets and distributed on the available nodes 
\cite{salloum}. 
In this paper, we propose a model for securing accuracy
in datasets present in EC nodes acting proactively and rejecting 
any data that could jeopardize their `solidity'.
We consider a Machine Learning (ML) algorithm
that decides if the incoming data should be stored locally or offloaded 
in peer nodes/Cloud.
Actually, we propose the use of Naive Bayesian Classifier (NBC)
for getting the final decision.
However, this decision is made over only features that 
are judged as significant for each dataset.
We consider that the remaining features should not be part of the 
decision making as they do not exhibit 
the appropriate and necessary characteristics that will lead to 
efficient analytics.

\textbf{\textit{Motivating Example}}. Feature selection models are widely adopted to filter irrelevant or redundant features in our datasets. It is a significant technique 
that is, usually, incorporated in dimensionality reduction models to deal with the 
so-called curse of dimensionality. 
In general, it always helps analyzing the data up front and, then, we are ready 
to support any decision making process.
Instead of collecting the data and performing any pre-processing/analysis action afterwards,
it would be better to make the analysis during their collection.
Hence, data quality and preparation can be secured before the dataset be the subject of any processing activity.
This process can become the groundwork for the subsequent engineering steps providing a solid foundation for building good ML schemes for decision making. 
When solid datasets are the final outcome, we can easily deliver analytics based on the specific features detected during the reception of data. Hence, no need for post-processing is present while the accuracy of data are at a high level.

Our intention is to provide a decision making model for securing data quality based on an 
ML scheme that will produce 
the relevant knowledge about the domain relationships during the reception of data.
A set of research efforts 
focus on the data quality management and they 
have identified its 
necessity in any application domain. 
However, they seldom discuss how to effectively validate data
to ensure data quality
\cite{refgao}.
The poor quality of data
could increase costs and reduce the efficiency 
of decision making \cite{nelson}.
In IoT, it is often necessary to detect correlations between
the collected data and external factors. 
We propose to secure data quality by allocating them to the appropriate datasets 
and select beforehand a (sub-)set of features that can be adopted in 
interpretable/explainable ML schemes.
Explainable models can be easily `absorbed' by humans depicting the
hidden correlations between data and giving the necessary 
insights to understand the reasons behind the adoption of the 
specific ML model.
The decision of the data allocation is performed over the selected features
to have the delivered datasets ready to be processed by the desired ML models.
Instead of performing the feature selection process after the collection of data,
we go a step forward and propose the execution of the activity during the 
reception of data.
Evidently, feature selection and data allocation are utilized at the same time to secure quality 
over a streaming environment. With this approach, we can save time and resources compared to a scheme where a batch processing activity is realized.

We build on an ensemble scheme, i.e., we adopt three (3) 
different model-agnostic approaches:
the Permutation Feature Importance (PFI) \cite{breiman}, 
Shapley Values and the 
Feature Interaction Technique (FIT)
\cite{friedman}. 
In addition, for delivering the final significance value for each 
feature through an aggregation of the three aforementioned outcomes,
we adopt an Artificial Neural Network (ANN) \cite{refalpaydin}.
ANNs do not represented explainable models but, in our case, 
the adopted inputs are the outputs of the aforementioned interpretable models.
The ANN undertakes the responsibility of `aggregating' the opinion of 
`experts' (i.e., our interpretable models) and deliver the final outcome.
Based on these technologies, we are able to detect the most significant features in the collected data
and build a powerful scheme for securing the data quality at the edge of the network.
We depart from legacy solutions and 
instead of collecting huge volumes of data and 
post-process them trying to derive knowledge, 
we propose their real time management and allocation 
keeping similar data to the same partitions.
The difference from our previous work presented in 
\cite{kolomvatsoscomputing} is that the current work
proposes an interpretable ML approach to 
give meaning to the stored data and the results
as delivered by the processing that end users desire.
The following list reports on the advantages 
of the proposed model: (i) we proactively `prepare' the data before the actual processing is applied;
(ii) we offer an interpretable ML scheme for satisfying the meaningful knowledge extraction;
(iii) we provide an ensemble scheme for aggregating multiple interpretable ML
models; (iv) we offer an ANN for delivering the most significant features fully aligned 
with the collected data; (v) the proposed model proactively secures the quality of data as it excludes data that may lead to an increased error;
(vi) our scheme leads to the minimum overlapping of the available datasets that is the target of the legacy data separation algorithms. 

The rest of the paper is organized as follows.
Section \ref{section2} reports on the related work while Section
\ref{section3} presents the problme under consideration.
In Section \ref{section4}, we present the adopted interpretable ML models and 
our ensemble scheme for combining the provided outcomes.
In Section \ref{section5}, we perform an extensive evaluation assessment and Section \ref{section6}
concludes our paper by giving insights in our future research plans.

\section{Related Work}
\label{section2}
The interested reader can find a survey
of data quality dimensions 
in \cite{sidi}.
Data mining and statistical techniques can be combined 
to 
extract the correlation of data quality dimensions, thus,
assisting in the definition of a holistic framework.
The advent of large-scale 
datasets as exposed by IoT
define additional requirements on 
data quality assessment.
Given the range of 
big data
applications, 
potential consequences of bad data quality can be 
more disastrous and widespread \cite{rao}.
In \cite{merino}, the authors 
propose the `3As Data Quality-in-Use model' composed 
of three data quality characteristics i.e., 
contextual, operational and 
temporal adequacy. 
The proposed model could be
incorporated in any large scale data framework
as it is not dependent 
on any technology. 
A view on the 
data quality issues in
big data
is presented in \cite{rao}.
A survey on data quality assessment methods is discussed
in \cite{cai}. 
Apart 
from 
that,
the authors present an analysis
of the data characteristics in large scale data environments 
and describe the quality challenges.
The evolution of the data quality issues in large scale systems is
the subject of \cite{batini}.
The authors 
discuss various relations between data quality 
and multiple research requirements.
Some examples are: the variety of data types, 
data sources and application domains, 
sensor networks and official statistics. 

ML interpretability is significant to deliver models that 
can explainable to humans, thus, to support efficient decision making.
There are varying definitions of it \cite{doshi}, \cite{lipton} without 
having a common ground, e.g., 
no formal ontology of interpretability types.
However, in \cite{lipton} is argued that these types can generally
be categorised in (i) transparency
(direct evidence of how the internals of a model work); 
or (ii) post hoc explanation (adoption of mapping methods to visualize input features
that affect outputs) \cite{montavon}, \cite{ribeiro}. 
A common
post hoc technique incorporates explanations by
example, e.g., case-based reasoning approach to select
an appropriately-similar example from training set \cite{caruana} or 
natural language explanations \cite{hendricks}.
The emergence
of these methods shows there is no consensus on how to assess the explanation quality \cite{carvalho}.
For instance, we have to decide the most appropriate metrics to assess the quality of an explanation.
Especially, for edge computing such issues are critical; the interested reader can find a relevant survey
of major research efforts where ML has been
deployed at the edge of computer networks in
\cite{murshed}.

In \cite{yazizi}, the authors discuss
the feasibility of running
ML algorithms, both training and inference, on a Raspberry Pi, an embedded version
of the Android operating system designed for IoT device development. 
The focus is to reveal the performance of various algorithms 
(e.g., Random Forests, Support Vector Machines, Multi-Layer Perceptron)
in constrained devices.
It is known that the highly regarded programming libraries
consume to much resources to be ported to the embedded processors
\cite{szydlo}.
In \cite{preece}, a service-provisioning
framework for coalition operations is extended to address
specific requirements for robustness and interpretability, allowing
automatic selection of service bundles for intelligence, surveillance
and reconnaissance tasks. 
The authors of \cite{roscher} review explainable machine learning
in view of applications in the natural sciences and discuss three core elements
i.e., transparency, interpretability, and explainability. 
An analysis of the convergence rate of an ML
model is presented in \cite{wang}.
The authors focus on a
distributed gradient descent scheme from a theoretical point of view
and propose a control algorithm that determines
the best trade-off between local update and global parameter
aggregation.

The `combination' between EC and deep learning is discussed 
in \cite{han}.
Application scenarios for both are presented together with 
practical
implementation methods and enabling technologies.
Deep learning models have been proven to be an efficient
solution to the most complex engineering challenges while at the
same time, human centered computing in fog and mobile
edge networks is one of the serious concerns now-a-days \cite{gupta}.
In \cite{puri}, the authors present a model
that learns a set of rules to globally explain the behavior of black box ML models.
Significant conditions are firstly extracted being evolved 
based on a genetic algorithm.
In \cite{liu}, an approach for image
recognition having the process split into two layers is presented.
In \cite{lane}, the authors
present a software accelerator that enhances deep learning execution on heterogeneous hardware.
In \cite{shamili} the authors propose the utilization of a Support Vector Machine (SVM) running on networked mobile devices to detect malware. A generic
survey on employing networked mobile devices for edge computing is
presented in \cite{tran}.
A combination of ML
with Semantic Web technologies in the context of model explainability is discussed in 
\cite{seeliger}. The aim is to semantically annotate parts of the ML models and 
offer the room for performing advanced reasoning delivering knowledge.
All the above efforts aim at supporting the Explainable 
Artificial Intelligence (XAI) \cite{lecue}.
XAI will facilitate industry to apply AI in products at scale, 
particularly for industries operating with critical systems.
Hence, end users will, finally, be able to enjoy 
high quality services and applications. 

\section{Problem Definition}
\label{section3}
Consider a set of $N$ edge nodes connected with a number of IoT devices.
IoT devices interact with their environment and collect data while being capable of 
performing simple processing activities.
Data are transferred in an upwards direction towards the Cloud infrastructure where
they are stored for further processing.
As exposed by the research community \cite{pham}, processing at the Cloud faces
increased latency compared to the processing at the edge of the network.
Therefore, edge nodes can maintain local datasets that can be the subject of 
the desired processing activities close to end users.
In each local dataset 
$D_{l}, l =1, 2, \ldots, N$, 
an amount of data (tuples/vectors) are stored. 
We focus on a multivariate scenario, i.e.,
$D_{l}$'s contain vectors in the form
$\mathbf{x} = \left\langle x_{1}, x_{2}, \ldots, x_{M} \right\rangle$
where $M$ is the number of dimensions/features.
Without loss of generality, we consider the same 
number of features in every local dataset.

The upcoming intelligent edge mesh \cite{sahni} incorporates
the necessary intelligence to have the edge node acting autonomously when
serving end users or applications. 
This way, we can deliver the desired services in real time fully aligned with the 
needs of end users/applications and the available data.
Arguably, the intelligent edge mesh provides 
analytics capabilities over the collected contextual data, thus,
edge nodes should conclude ML models that have meaning for end users/applications.
For instance, edge nodes may perform ML models 
for novelty or anomaly detection.
When delivering ML models, a challenging problem is to extract higher-valued features that 
`represent' the local dataset, thus, we can get our strategic decisions only over them
and deal with the so-called curse of dimensionality.
Formally, we want to detect the most significant features $x_{ij}, j = 1, 2, \ldots, M$ 
based on the available data vectors
$\mathbf{x}_{i}, i = 1, 2, \ldots, |D_{l}|$.
Hence, we will be able to `explain' the local ML model 
making end users/ applications to have faith in it.
This is the main motivation 
behind the adoption of ML model interpretability.
We have to notice that the selected features are those: (i) being the most significant 
for each dataset, thus, they have to be part of any upcoming processing; (ii) being adopted to secure data quality by incorporating them in the decision for the allocation of the incoming data to the appropriate datasets; (iii) being the most appropriate to support the explainability of the subsequent ML schemes. 

Local datasets are characterized by specific statistical information,
e.g., mean and variation/standard deviation.
The aim of each node is to keep the accuracy of the local dataset at high levels.
The accuracy is affected by the error
between $D$ and $\mathbf{x}$.
Edge nodes should decide if 
$\mathbf{x}$ `matches' $D$, however,
based on features that are detected as significant for the
local dataset (and not all of them).
Through this approach, we do not take into consideration 
features that are not important for the local ML model
as exposed by the incoming data vectors.
We perform a dimensionality reduction beforehand during the collection of data.
This means that our scheme is fully aligned with the needs of the 
environment (where edge nodes and IoT devices act) and 
end users/applications.
If $\mathbf{x}$ deviates from $D$, it can `rejected' and transferred either 
in a peer node (where it exhibits a high similarity)
or in Cloud (as proposed in \cite{kolomvatsoscomputing});
its incorporation in $D$ will affect the local statistics
`imposing' severe fluctuations in 
basic statistical measures (e.g., mean, deviation).
A Naive Bayesian Classifier (NBC) is adopted to  
deliver the decision of locally storing 
$\mathbf{x}$ or offloading it in peers/Cloud.
The NBC reports over the probability of having $\mathbf{x}$
`generated' by the local dataset $D$.
However, the decision is made over the most significant 
features as delivered by the proposed 
ensemble interpretable ML model aiming at 
having an ML model that can be explained 
in end users/applications.
Our ensemble scheme involves three
interpretable, model agnostic techniques, i.e., 
the PFI, 
Shapley Values and the 
FIT.

\begin{figure}[h]
\centering
\includegraphics[width=0.5\textwidth]{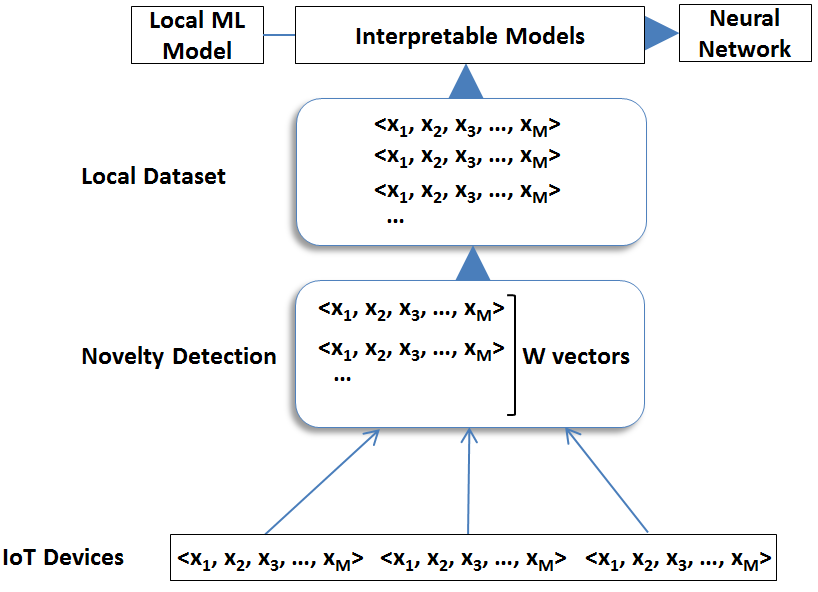}
\caption{The architecture of an edge node.}
\label{fig1}
\end{figure}

For handling the `natural' evolution of data in the error identification
(between $D$ and $\mathbf{x}$),
we consider a novelty detection model before the incoming data being subject of the
envisioned NBC (for deciding the storage locally or the offloading
to peers/Cloud).
The novelty detection is applied over a copy of the latest $W$ vectors and
delivers if there is a significant update in the statistics of the incoming data.
When the novelty detection module identifies 
the discussed update, the $W$ data vectors are incorporated in the local dataset 
$D$ and the proposed interpretable ML model is fired.
In this paper, due to space limitations, we do not focus on a specific
novelty detection scheme and consider 
a indicator function $I(\left[\mathbf{x}\right]^{W}) \to \left\lbrace 0, 1, \right\rbrace$
to depict the change in the incoming data statistics.
For achieving the `final' interpretability,
we propose the use of an ANN over 
multiple model-agnostic interpretable models. 
The goal is to decouple the model from the interpretation 
paying more attention on the significance of
each feature and the amount of its contribution in the `black box' ML model
(i.e., the NBC).
The ANN receives as inputs the 
outcomes of each interpretable technique and deliver the final 
value to decide over the features that are significant for the 
local dataset. 
In any case, even if ANNs are not interpretable models, the 
interpretability in our approach is secured by the three aforementioned explainable 
schemes.
The ANN is adopted to `aggregate' the `opinion' of three different interpretable models
and get the final outcome based on which we, consequently, get the significance of a feature.
The ANN is there to handle possible `disagreements' for the 
the significance of each feature.
In Figure \ref{fig1}, we can see the envisioned setup.
In the first place of our future research plans is 
the aggregation of interpretable models 
originated in different edge nodes to deliver 
and interpretable model for a group of nodes covering a specific
area.

\section{The Ensemble Scheme}
\label{section4}

\subsection{Feature Effects \& Selection}
An NBC adopts the Bayes theorem of conditional probabilities to estimate 
the probability for a class given 
the value of the feature.
This is realized for each feature independently;
a similar approach as having an assumption of the independence of 
features.
Given a dataset $X$ and its values $[x_{i}]$, the probability of 
a class $C_{k}$ is given by:
\begin{equation}
 P\left( C_{k} | X \right) = \frac{1}{Q} P(C_{k}) \prod_{i=1}^{n} P(x_{i}|C_{k})
\end{equation}
where $Q$ is a scaling parameter adopted to secure that probabilities 
for all the classes sum up to unity.
The independence assumption leads to an interpretable model,
i.e., for each classification, its contribution to the predicted class
is easily perceived. 

Let the dataset be $\mathbf{Z} \ \left[ y, X \right]$ where $y$ 
is the output $c$-length vector and a 
$cXp$ covariate matrix.
In addition, we get the trained model 
$f$ over our dataset and the 
$L(y,f)$ is a function delivering the
error measure for our model based on the 
outcome $y$.
The PFI scheme \cite{fisher} adopts a number of steps
for calculating each feature's importance
to finally decide the final (sub-)set of 
the adopted features.
The training dataset is split in half and 
values of the $j$th feature are swapped 
between the two hales instead of producing permutation for the
feature. 
Initially, the model 
estimates the $f$'s error notated 
as $e^{o} = L(y, f(X))$ based on any technique
(e.g., we can adopt the mean squared error).
Afterwards, for each feature, we generate feature permutations in data
breaking the correlation between the feature and the outcome $y$.
For this permutation, we calculate the 
error
$e^{p} = L(y, f(X^{p}))$
where $X^{p}$ is the dataset delivered
after the permutation.
The PFI for the feature is calculated as follows:
$F_{j}^{PFI} = \frac{e^{p}}{e^{o}}$.

Shapley values are originated in the coalition game theory.
The interpretation of a Shapley value 
$\xi_{ij}$ for the feature $j$ and the instance $i$ of the dataset 
is the feature value $x_{ij}$ contributed $\xi_{ij}$ towards the estimation 
for $i$ compared to the average prediction 
for the dataset.
A Shapley value aims at detecting the effect of the $j$th feature on the prediction 
of a data point.
For instance, in a linear model,
i.e.,
$\hat{f}(x_{i})=\beta_{0}+\beta_{1}x_{i1} + \beta_{2}x_{i2} + \ldots + \beta_{p}x_{ip}$, it is easy through the 
weight $\beta_{j}$ to expose the 
effect of the $j$th feature.
For retrieving the final Shapley value, we should examine all possible `coalitions' of features which a computational intensive task
when we focus on a high number of features.
In these coalitions, we have to incorporate or
leave the feature in combination with other features to see its effect in the estimation of the target parameter.
Hence, we rely on an approximation model
proposed in 
\cite{strumbel}.
The method is based on a Monte-Carlo simulation that delivers the final 
value, i.e.,
$F_{j}^{SV} = \frac{1}{M} \sum_{m=1}^{M} ]\left( \hat{f}(x^{+j}) - \hat{f}(x^{-j})\right)$.
In this equation,
$M$ is the number of iterations (we get the mean of the differences),
$\hat{f}$ is the estimated value for the $i$th sample  
based on the black box ML model,
$x^{+j}$ is the selected instance with 
a random number of features replaced 
by values retrieved by a random data point $x$ and
$x^{-j}$ is identical to $x^{+j}$ but we exclude the $j$th feature.
This means that we create two new instances
$x^{+j}$ \& $x^{-j}$ from the same dataset, however, performing a sampling for 
realizing features permutations.
The steps of the approach are as follows:
(i) select an instance of interest $i$ and a feature $j$;
(ii) select the number of samples $M$;
(iii) for each sample, select a random instance and mix the order of features;
(iv) create two new instances (as described above) for the $i$th sample;
(v) get the difference of the estimated value;
(vi) get the mean of the results as the final Shapley value.

We can estimate the FIT value for each feature based on the so-called
Partial Dependence (PD) between
features.
The interaction of a feature with all the remaining in our model will depict the
significance of the specific feature.
Let two features $x_{}j$ and $x_{k}$.
For measuring if the $j$th features interacts with the remaining features in the model, we get:
$F_{j}^{FIT} = \frac{\sum_{i=1}^{n} \left[ \hat{f}(x^{(i)}) - PD_{j}(x_{j}^{(i)})-PD_{-j}(x_{-j}^{(i)}) \right]}{\sum_{i=1}^{n}}$
($-j$ represents the exclusion of the $j$ feature from the instance).
The partial function for a feature can be easily retrieved by 
a Monte Carlo simulation,
i.e., 
$PD(x_{j}) = \frac{1}{n} \sum_{i=1}^{n} \hat{f}(x_{j}, \dot{x})$
where
$\dot{x}$ are values from the dataset for features we are not interested in. 

\subsection{Combination of Multiple Models}
The combination of the interpretable models is performed
for each feature through the use of our ANN.
ANNs are computational models inspired by natural neurons. 
The proposed ANN is a series of functional 
transformations involving $C$ combinations of the input values 
i.e., 
$o_{f}^{1}, o_{f}^{2}, \ldots, o_{f}^{|\mathcal{O}|}$ ($o_{f}^{k}, k = 1, 2, \ldots, |\mathcal{O}|$ 
($o_{f}^{k}$ is the final fused value for each metric) \cite{refbishop}.
The linear combination of inputs has
the following form:
$\alpha_{j} = \sum_{k=1}^{|\mathcal{O}|} w_{jk} o_{f}^{k} + w_{j0}$,
where $j = 1, 2, \ldots, C$. 
In the above equation, $w_{jk}$ are weights and $w_{j0}$
are the biases.
Activation parameters $\alpha_{j}$ are, then, transformed by adopting 
a nonlinear activation function to give 
$z_{j} = g(\alpha_{j})$.
In our model, $g(.)$ is the sigmoid function.
The overall ANN function is given by:
\begin{eqnarray}
y(\mathbf{o_{f}}) = s\left( \sum_{j=1}^{C} w_{j} g\left( \sum_{k=1}^{|\mathcal{O}|} w_{jk} o_{f}^{k} + w_{j0} \right) + w_{0} \right),
\label{eq:d2}
\end{eqnarray} 
where $s(.)$ is the sigmoid function defined as follows: 
$s\left( \alpha \right) = \frac{1}{1+exp(-\alpha)}$.
In addition, 
$C$ is the combinations of the input values and $\mathcal{M}$
is the number of the inputs.

The proposed ANN tries to aggregate heterogeneous metrics and 
pay attention on their importance. 
We adopt a three layered ANN. 
The first layer is the \textit{input layer}, 
the second is the \textit{hidden layer} and 
the third is the \textit{output layer}. 
We adopt a feed forward ANN 
where data flow from the input layer to the output layer. 
In our ANN, there are $|\mathcal{O}|$ inputs 
i.e., the final estimated values for each performance metric
depicted by the vector $\mathbf{o_{f}}$. 
The output \textit{$y(\mathbf{o_{f}})$} 
is the aggregated value that will be the basis for deciding 
the significance of each feature.
Actually, we fire the ANN and get the significance value of each feature creating, at the end,
a sorted list.
We adopt a threshold $d$ above which a feature is considered as
significant for our model.
The most important part of our decision scheme 
is the training of the proposed ANN. 
In the training phase, 
we adopt a training dataset depicting various strategies / scenarios 
concerning the interpretable ML models.
This training dataset contains various combinations of outcomes 
of the adopted interpretable models. 
For a number of iterations, 
we produce values that correspond to multiple combinations of metrics 
depicting various states of the network and the node.
The dataset is defined by experts.

\section{Performance Assessment}
\label{section5}

\subsection{Indicators \& Simulation Setup}
We present the experimental evaluation of the proposed model through a set of simulations. It is worth noticing that our simulator was developed in R and our experiments were performed using the dataset provided by \cite{Zheng}. The discussed dataset relates with real-world QoS evaluation reports by 339 users on 8,525 Web Services. 
Our evaluation focuses on the improvement of the decision-making process when deciding whether to keep data locally based on the most 
important features of the incoming data as opposed to all of them, 
i.e., no interpretability (feature selection) process is applied. Furthermore, we are concerned with keeping locally the instances of data that preserve the solidity of the current dataset maintained by an EC node. Solidity is very important as it can be used to enhance the confidence interval of the statistics information of datasets. In our experimental evaluation, we do not pay attention 
on the specific features that are selected in every evaluation scenario. 
The ultimate goal is to detect if the final outcome corresponds to something valid and interesting from the application point of view (i.e., secure quality by allocating data to the appropriate datasets).
Lastly, we focus on the time required for a node to make a decision. 

We define the metric $\Delta$ as the percentage of correct decisions that are made. The following equation holds true: $\Delta=|CD|/|D|*100$\%. In the aforementioned equation, $CD$ represents the set of correct decisions related to the storage of the appropriate data locally and $D$ represents the set of decisions taken in our experimental evaluation. When $\Delta \to 100$\%, it means that the model has a high accuracy, whereas as $\Delta \to 0$\%, the model's predictions are not reliable at all. Moreover, we establish the metric $\sigma$, which is depicted by the standard deviation of data and describes the `solidity' of the local dataset. The lower the $\sigma$ becomes, the more `solid' a node's dataset is and the opposite is true when $\sigma$'s value becomes high; specifically, when a dataset is quite `solid', it means that its values are concentrated around the mean value, hence, giving us a concrete idea of the concentration of data. Having a `solid' dataset can be highly useful in the efficient allocation of queries to datasets that can serve them in the most effective manner. In addition, we report on $\tau$, representing the average time that is required for a decision to be made on whether a single data instance should be kept locally, or offloaded to another EC node into which it fits better or the Fog/Cloud.
	
We perform a set of experiments for a variety of $M$ and $w$ values. We adopt $M \in \left\lbrace 10,50,100 \right\rbrace$, i.e. different numbers of dimensions for the dataset, as well as $w \left\lbrace 10\%,20\%,50\% \right\rbrace$, i.e., different percentages of features to be used for the final decision about a data instance's storage node. 

\subsection{Experimental Outcomes}
We start by evaluating our model in terms of $\Delta$ (see Figure 2). In this set of experiments, we compare the performance of two models, i.e., CD and wCD. The former depicts the percentage of correct decisions made by the Naïve Bayes Classifier based on all the features of the adopted dataset. This is a baseline solution where equal significance is paid for all the available features. The latter model illustrates the percentage of correct decisions made by the Naïve Bayes Classifier based only on the $w$*$M$ most significant features of the dataset. It consists of the model where our `reasoning' is adopted to detect the most important features of the dataset. We observe that in the majority of the experimental scenarios (except one case), the performance of wCD is decidedly improved when compared against the CD. This is quite logical as in wCD the Naïve Bayes Classifier is able to focus solely on the most important features of an instance to make a decision about whether to keep it locally or not and does not take into account features that can result in a false prediction. This provides an evidence that our mechanism is capable of efficiently detecting significant features, thus, we can adopt them to support decision making. As $M$ increases, $\Delta$ becomes low, since an increment in the number of features used by the classifier brings about the aforementioned false predictions. Features that are not significant steer the prediction away from the actual class, and even if only $w$*$M$ of the features are used, the features are still too many to make the decision-making process as clear as it needs to be.
In general, the performance of the proposed system is affected by $M$ and $w$, i.e., 
increased $M$ \& $w$ lead to lower $\Delta$ values.

\begin{figure}[h]
\centering
\includegraphics[width=0.5\textwidth]{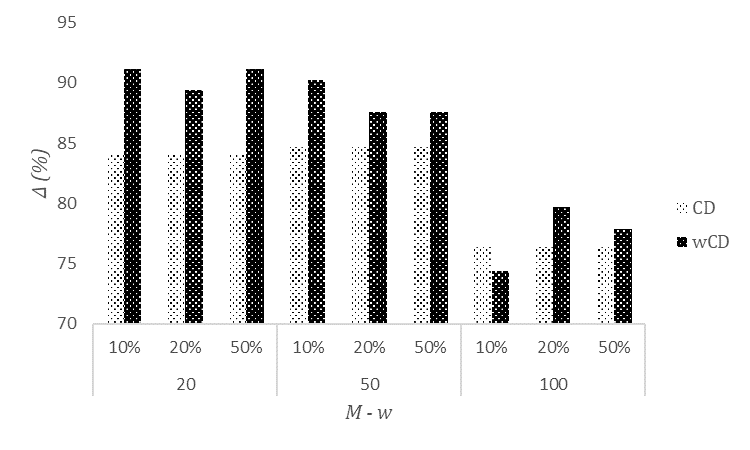}
\caption{Performance evaluation for the correct decisions derived by our model.}
\label{fig2}
\end{figure}

In Figure 3, we present our results for the solidity of the retrieved datasets after the selection of the most significant features. 
In this set of experiments, we compare three models, i.e., the OS, the BNS and the NNS.
OS represents the model where we deliver the 
$\sigma$ realization based on the entire set of data available in a node.
The BNS depicts the solidity of the dataset when adopting the Naïve Bayes Classifier and the entire set of the available features.
Finally, the NNS represents the solidity of the dataset when adopting the features selected by the proposed interpretable approach.
In all the experimental scenarios, our feature selection approach (i.e., the NNS) manages to achieve the best performance.
This means that the final, delivered dataset is solid and the deviation from the mean is limited. 
Hence, we can increase the accuracy of data as 
they do not deviate from the mean limiting the possibilities of the presence of extreme values that can negatively affect the statistical characteristics of the dataset.
Apart from that, in a latter step, we can 
create data synopses to be distributed in the upper layer of a Cloud-Edge-IoT architecture that could be characterized by an increased confidence interval.
In Figure 3, we also observe that the OS exhibits the worst performance among the compared models.
Finally, a low $M$ combined with a low $w$ leads to best possible performance.

\begin{figure}[h]
\centering
\includegraphics[width=0.5\textwidth]{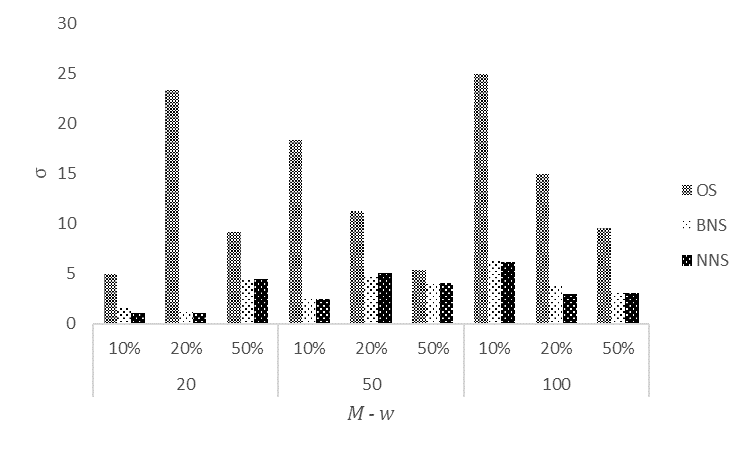}
\caption{Data solidity as delivered by the proposed model.}
\label{fig3}
\end{figure}

The last set of our experiments deal with the time required to conclude the final sub-set of features. In Figure 4, we plot $\tau$ for various combinations of $M$ and $w$.
We have to notice that $\tau$ is retrieved as the mean for a number of iterations. As it can be observed, $w$'s increment does not reflect any change to $\tau$. This is reasonable since, the model has to do calculations for each of the $M$ features to determine the most important ones. This procedure is repeated for each instance and its total duration is higher than the decision itself. Figure 4 also depicts that $\tau$ is (approx.) linear to the total number of features $M$.
This observation becomes the evidence of the efficiency of the proposed approach as it `transparent' the total number of features taken into consideration.

\begin{figure}[h]
\centering
\includegraphics[width=0.5\textwidth]{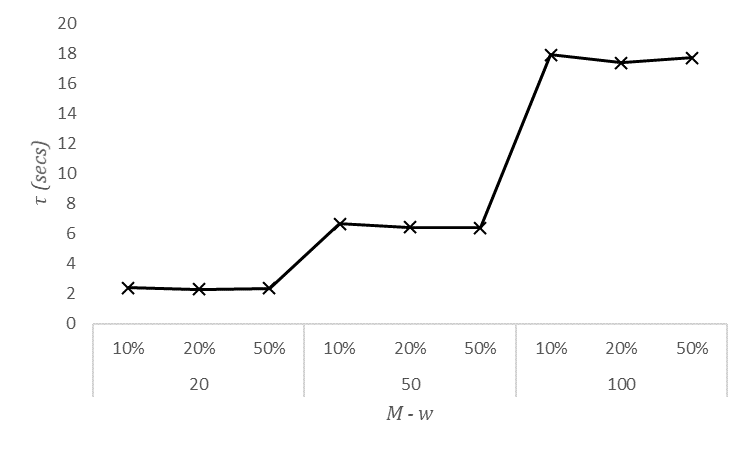}
\caption{Performance evaluation related to $\tau$.}
\label{fig4}
\end{figure}

\section{Conclusions \& Future Work}
\label{section6}
Data quality is significant because without it, we are not able to support efficient decision making. 
Securing data quality will give a competitive advantage 
especially to companies that are based on 
various analytics processing activities. 
In this paper, we focus on the management of data quality and propose that any decision related to the acceptance of
incoming data should be based on specific features and not all of them.
Such features will exhibit the appropriate statistical characteristics that will make, afterwards, the desired analytics explainable to end users.
We assume an edge computing environment and 
propose and ensemble scheme for features selection.
We present the adopted algorithms and provide the 
aggregation process.
In addition, we propose the use of a Neural Network that 
delivers the importance of each individual feature before we conclude the final sub-set.
Based on the above, 
we are able to detect the most significant features 
for data present at edge nodes. 
Our experimental evaluation exhibits the performance of the system and its capability to select the proper features.
Our numerical results denote the significance of our model and its capability to be adopted in real time applications.
In the first place of our future research plans, we will provide a mechanism for covering the uncertainty around the significance of each feature. Additionally, we plan to incorporate into our model a scheme that delivers the selection decision based on a modeling of the available features adopting a sliding window approach. 

\section*{Acknowledgment}
This research received funding from the European's Union Horizon 2020 research and innovation programme under the grant agreement No. 745829.


\end{document}